\documentclass[runningheads]{llncs}
\usepackage{graphicx}
\usepackage{caption}
\usepackage{authblk}
\usepackage{subcaption}
\usepackage[section]{placeins}
\usepackage{hyperref}
\usepackage{MnSymbol}
\usepackage[backend=biber,style=ieee]{biblatex}
\usepackage{lscape} 
\usepackage{adjustbox}
\usepackage[nottoc]{tocbibind}
\usepackage{setspace}
\usepackage{algorithm}
\usepackage{algpseudocode}
\usepackage{enumitem}
\usepackage{makecell}
\usepackage{color}
\usepackage{tikz}
\usepackage{wrapfig}
\usepackage{rotating}
\usepackage{adjustbox}
\usepackage{afterpage,lscape}
\usepackage{fontawesome}
\usetikzlibrary{shapes}

\graphicspath{{Fig/}}

\begin{document}
\title{Concept Drift Detection using Ensemble of Integrally Private Models}
%
%
\author{Ayush K. Varshney\inst{1}\orcidID{0000-0002-8073-6784} \and
Vicen\c{c} Torra\inst{1}\orcidID{0000-0002-0368-8037}}
\authorrunning{A. K. Varshney et al.}
%
\institute{Department of Computing Sciences \\ Umeå University 90740, Sweden\\
\email{\{ayushkv,vtorra\}@cs.umu.se}}
\maketitle              
\begin{abstract}

Deep neural networks (DNNs) are one of the most widely used machine learning algorithm. DNNs requires the training data to be available beforehand with true labels. This is not feasible for many real-world problems where data arrives in the streaming form and acquisition of true labels are scarce and expensive. In the literature, not much focus has been given to the privacy prospect of the streaming data, where data may change its distribution frequently. These concept drifts must be detected privately in order to avoid any disclosure risk from DNNs. Existing privacy models use concept drift detection schemes such ADWIN, KSWIN to detect the drifts. In this paper, we focus on the notion of integrally private DNNs to detect concept drifts. Integrally private DNNs are the models which recur frequently from different datasets. Based on this, we introduce an ensemble methodology which we call 'Integrally Private Drift Detection' (IPDD) method to detect concept drift from private models. Our IPDD method does not require labels to detect drift but assumes true labels are available once the drift has been detected. We have experimented with binary and multi-class synthetic and real-world data. Our experimental results show that our methodology can privately detect concept drift, has comparable utility (even better in some cases) with ADWIN and outperforms utility from different levels of differentially private models. The source code for the paper is available \hyperlink{https://github.com/Ayush-Umu/Concept-drift-detection-Using-Integrally-private-models}{here}. \footnote{Accepted for publication in MLCS co-located with ECML-PKDD 2023.}

\keywords{ Data privacy \and Integral privacy \and Concept Drift \and Private drift \and Deep neural networks \and Streaming data.}
\end{abstract}

\section{Introduction}

\begin{figure}
    \centering
    \includegraphics[width=0.65\textwidth]{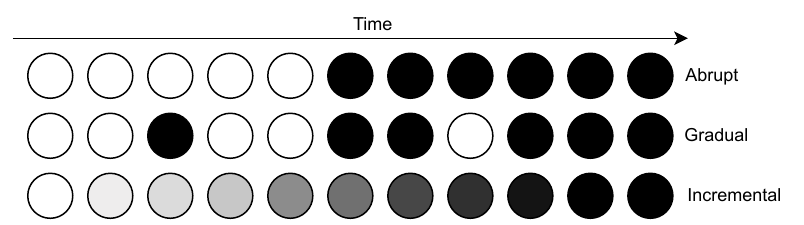}
    \caption{Types of drifts in the data}
    \label{types of drifts}
\end{figure}

In recent years, the interest in deep learning models has witnessed a steady increase, despite encountering various challenges such as explainability, privacy, and data dependency. To address these issues, significant advancements have been made, including approaches to enhance explainability \cite{schwab2019cxplain}, privacy-preserving techniques \cite{shokri2015privacy}, adopting a data-centric perspective to facilitate model training with high-quality data, and more. However, limited attention has been given in the context of streaming data, which refers to the continuous arrival of data in real-world scenarios, often accompanied by the problem of concept drift. Concept drift implies that the statistical properties of the data may change over time, necessitating the model to adapt to these changes to ensure reliable predictions. Noisy data at one point of time may become useful data over time. These changes in data distributions can be due to various hidden factors. Handling of such drifts is a must and has been employed in many applications such as spam detection \cite{gama2004learning}, demand prediction \cite{vzliobaite2016overview}. Learned models must have the ability to detect concept drifts and incorporate them by retraining on the new data. Three types of concept drifts have been shown in Fig. \ref{types of drifts}. Abrupt drifts are sudden changes in the data distribution. E.g. complete lockdown in many countries due to COVID-19 pandemic. Gradual drifts are the drifts which changes the distribution over time. E.g. in fraud detection system, fraudsters adapt according to the improving security policies in place. Incremental concept drift are the drifts where old concepts vanishes completely with time. E.g. after lifting COVID-19 lockdown, people may be hesitant to return to their normal behaviour. 

In the literature of concept drift detection, there has been several algorithms which can detect concept drifts such as Adaptive windowing (ADWIN) \cite{bifet2007learning} and its variant, and Kolmogorov-Smirnov Windowing (KSWIN). These are the two prominent drift detection  methods used in the streaming settings. To detect drifts, these techniques were originally proposed assuming the availability of true labels which is unrealistic in most real-world assumptions. ADWIN employs two windows, one fixed size and one variable size, which slide over the incoming data stream. The fixed size window keeps the most recent points and the variable size window keeps the earlier points. If the statistics of the two windows differs significantly then ADWIN indicates that the drift has been detected.

In case of DNNs, training requires huge amount of data and acquisition of ground truth to detect drift can be very costly. A recently proposed uncertainty drift detection scheme \cite{baier2021detecting} detects drift during inference without the availability of true labels. It computes values for prediction uncertainty using dropout in the DNNs and uses entropy of these uncertainty values to detect drifts. Another approach to get prediction uncertainty is through the ensemble of DNN models. Different DNNs produce different probabilities during predictions and the overall uncertainty in their predictions can be used to detect the drift. In the literature, almost none of the approaches focus on the privacy perspective of drift detection.

Privacy is a crucial factor to take into account in concept drift detection as data is often sensitive. There exists many privacy models such as k-anonymity \cite{samarati2001protecting}, differential privacy (DP) \cite{10.1007/11787006_1}, integral privacy \cite{torra2020explaining} and others for static environment but their counter-parts for online learning are rather limited. For online learning, k-anonymity \cite{salas2018general} tries to protect against identity disclosure by guaranteeing k-anonymity for addition, deletion, and updating the records but may fail to protect attribute disclosure; differential privacy (DP) perturbs the data or the model in order to generate privacy-preserving outputs against the disclosure of sensitive information. Even though DP provides theoretically sound privacy-preserving models, it has a number of practical drawbacks. For instance, when aiming for high privacy (small $\epsilon$), the amount of noise added can become very high. Moreover, there is a finite privacy budget for multiple searches, and high sensitivity queries demand a bigger amount of noise. DP may struggle with the privacy budget when the data distribution changes frequently. You may end up loosing utility or privacy or both in the long run. Also, the addition of a lot of noise to the output can make machine learning models less useful. Most of the privacy approaches in the online learning literature focuses on either storing the data or predicting the output privately. None or very few approaches in literature focuses on detecting drifts privately.

In our approach, we have considered Integral Privacy as an alternative to DP to generate high utility, privacy-preserving machine learning models. Integrally-private models provide sound defence against membership inference attacks and model comparison attacks. A membership inference attack is about getting access to the records used in the training process. On similar lines, a model comparison attack gives intruder access to the complete training set or to a huge portion of the training set through intersectional analysis. A machine learning model is integrally private \cite{torra2020explaining} if it can be generated by multiple disjoint datasets. For an intruder whose aim is to do membership inference attacks or model comparison attacks, integrally-private models create ambiguity as the models are generated by multiple disjoint datasets. It has been 
proven in \cite{thudi2022necessity} that under some conditions it is possible to obtain, with probability close to one, the same parameter updates for a model with multiple minimatchs. They also find that a small fraction of a dataset can also lead to good results. One of the first works which shows the framework for model comparison attack and the defence by integral privacy for decision trees was given in \cite{senavirathne2019integrally}. The authors generate the complete model space and return the integrally private decision tree models which have approximately same model parameters. Generating complete (or approximately complete) model space can be a very computationally intensive task for a dataset with only few thousand instances. 

For DNNs, generating model space and comparing models to find integrally private models can be tricky. This is due to the fact that for a given layer of two different models, equivalent neurons can be placed in different positions. Also, due to huge number of learning parameters in DNNs, there can be very few recurring models. In order to overcome these challenges for DNNs, a relaxed variant of integral privacy, $\Delta$-Integral privacy, was proposed in \cite{varshney2023integrally}. $\Delta$-Integral privacy ($\Delta$-IP) considers models which are at most $\Delta$ distance apart, and then recommends the mean of these models (in the $\Delta$ range) as the integrally private model. The $\Delta$-IP algorithm can recommend up to X number of integrally private models which can be used as an ensemble of private models to detect concept drifts in streaming data. In this paper, we propose a methodology for drift detection through an ensemble of $\Delta$-integrally private models. We compute an ensemble of $\Delta$-IP models and use them to compute a measure of prediction uncertainty. This prediction uncertainty of $\Delta$-IP models on the incoming datastream is used to detect concept drift. Our methodology only requires true labels to recompute the $\Delta$-IP models once a drift has been detected. We also present the probabilistic analysis for the recurring models. Our theoretical analysis is inspired from the work in \cite{thudi2022necessity} which focuses on forging a minibatch. In our case, the analysis focus on learning similar parameters after complete training.

Our experimental setup shows results for ANN (one hidden layer with 10 neurons) and DNN of 3 hidden layers (10-20-10 hidden layer architecture). We evaluate our proposed methodology for 3 real-world dataset and 4 synthetic dataset. We have also compared our results with different levels of privacy in DP models. We show that our approach outperforms the DP alternatives. We find that ensemble of integrally private models can successfully detect concept drifts while maintaining the utility of non-private models. 

The rest of the paper is organized as follows. Section 2 describes the background for the proposed drift detection methodology. Section 3 describes our proposed work. Section 4 gives the experimental analysis. The paper finishes with some conclusions and future work.

\section{Background}

In this section we describe the major concepts that are needed in this work.

\subsection{Uncertainty in Neural Networks} 

Understanding the uncertainty of a model is essential to understand the model's confidence. In DNNs, class probabilities can not be the proxy for model's confidence. For unseen data, DNNs may give high probability even when the predictions are wrong. This can be the case in concept drifts i.e, the prediction may be uncertain but the system can give high class probability. Ensemble methods find the uncertainty using predictions from the family of DNNs. Here you train multiple DNNs with different initializations. In this way you generate a set of confidence parameters from multiple DNNs, and the variance of the output can be interpreted as the model uncertainty. In our work, we estimate the model uncertainty using an ensemble of private models. With drift in estimated uncertainty as an indicator for concept drift, we can employ drift detection schemes such as ADWIN, KSWIN to detect concept drift. 

\subsection{Model Comparison Attack and $\Delta$-Integral privacy}

Integral privacy \cite{torra2020explaining} is a privacy model which provides defense against model comparison attacks and membership inference attacks. In a model Comparison Attack \cite{senavirathne2019integrally, varshney2023integrally}, an intruder aims to get access to the sensitive information or do membership inference analysis by comparing the model parameters. A model comparison attack assumes that the intruder has access to the global model $M$ trained using algorithm $A$ on the training set $X$ a subset of the population $\mathcal{D}$, and some background information $S^* (\subseteq \mathcal{D})$. The intruder wants to get the maximum (or total) number of records used in the training process. I.e. the intruder wants to maximize access to $X$. The intruder draws a number of samples $S_1,S_2,...,S_n$ from $\mathcal{D}$ and compares the model generated by each $S_i$ with the global model $M$. Then, the intruder selects the $S_i$ corresponding to the most similar model to $M$ and hence guesses the records used in the training. In case that there are multiple models similar to $M$, the intruder can do intersectional analysis for membership inference. That is, find common data records which lead to the model $M$. In case of DNNs, model comparison can be tricky as highlighted before in \cite{varshney2023integrally}. The comparison between models is done by comparing each layer and neurons in respective layers. 

In order to defend against such attacks, integral privacy requires you to chose a model which recurs from different disjoint datasets. Disjoint datasets are needed to avoid intersectional analysis. In this way an intruder cannot identify the training set because multiple training sets lead to the same model. As explained in Section 1, due to the huge number of parameters in DNNs there are very few recurring models. $\Delta$-IP relaxes the equivalence relation between neurons. It allows the two models to be considered as equal if neurons in each layer of the respective model are at most $\Delta$ distance apart. Formally, $\Delta$-IP can be defined as follows.

\textbf{$\Delta$-Integral Privacy} Let $\mathcal{D}$ be the population, $S^* \subset \mathcal{D}$ be the background knowledge, and $M \subset \mathcal{M}$ be the model generated by an algorithm $A$ on an unknown dataset $X \subset D$. Then, let $Gen^*(M, S, \Delta)$ represent the set of all generators consistent with background knowledge $S^*$ and model $M$ or models at most $\Delta$ different. Then, k-anonymity $\Delta$-IP holds when $Gen^*(M, S, \Delta)$ has atleast k-elements and 
\begin{equation} \label{e1.1}
\bigcap_{S \in Gen^*(G, S^*, \Delta)} S = \emptyset
\end{equation} 

\section{Proposed Methodology}

\begin{figure}[!t]
    \centering
    \includegraphics[width=0.85\textwidth]{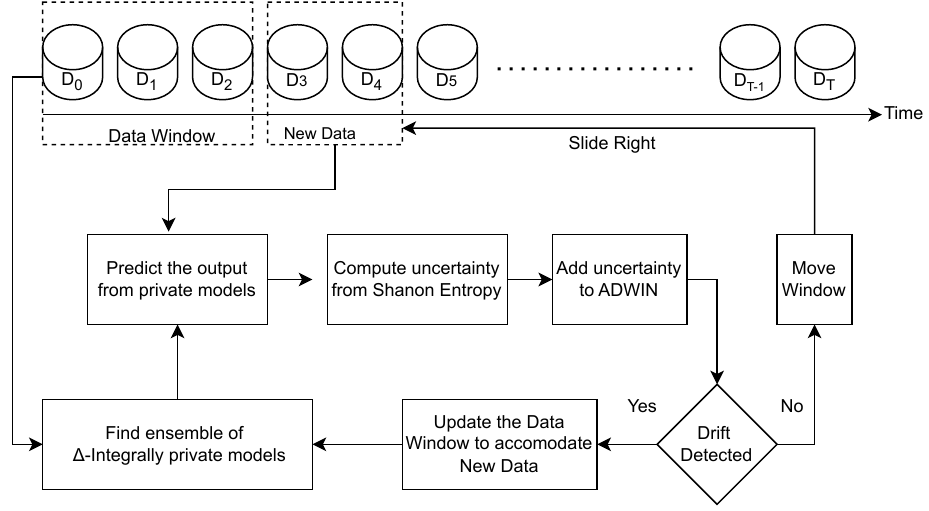}
    \caption{Flowchart drift detection using ensemble of $\Delta$-Integrally Private Models}
    \label{flowchart}
\end{figure}

In this section, we provide the details of our proposed methodology which we call \textit{Integrally private drift detection} (IPDD) scheme. Our proposed IPDD methodology detects drifts with unlabeled data but assumes that true labels are available on request. Our approach is based on the detection of concept drifts from the measure of uncertainty in prediction by ensemble of private models. Previous works \cite{kendall2017uncertainties} \cite{baier2021detecting} show that prediction uncertainty from DNN is correlated with prediction error. We argue on similar lines and use drift in prediction uncertainty as a proxy for detecting concept drift. We use Shanon entropy to evaluate the uncertainty over different $c$ class labels.

\begin{algorithm}[!ht]
\caption{Algorithm to generate k $\Delta$-Integrally private DNNs for training data $D$. The algorithm return an ensemble of k private models} \label{Algo IPDNNs}
\begin{algorithmic}
\State \textbf{Inputs:} $D$ - Training data  
\State \textbf{Output:} returns k integrally private models
\State \textbf{Algorithm:} \\
$N$ - Size of subsamples \\
$\Delta$ - Privacy parameter \\
S = Generate\_subsample($D$, $N$) \\ \Comment{Generate n disjoint subsamples of size $N$}\\
ModelList = [[]]
\While{ $S_i \leftarrow S$ } \Comment{For all samples in $S$}
    \State $M_i \leftarrow$ Train DNN on $S_i$
    \State present $\leftarrow$ False
    \If{$M_{i}$ is utmost $\Delta$ distance apart from models in ModelList} 
        \State Put $M_{i}$ in the same bucket
        \State present $\leftarrow$ True
    \EndIf
    \If{present is False}
        \State Append $M_i$ in ModelList \Comment{Create a new bucket with $M_i$}
    \EndIf
\EndWhile
\State \textbf{Returns} mean of top k recurring models from ModelList
\end{algorithmic}
\end{algorithm}

\begin{algorithm}[!ht]
\caption{Drift detection using $\Delta$-Integrally private models} \label{Algo IPDD}
\begin{algorithmic}
\State \textbf{Inputs:} $\mathcal{D}$ - Dataset
\State \textbf{Algorithm:} \\
training\_data = Initial\_data($\mathcal{D}$) \\ \Comment{Initial Data to train private DNNs}\\

Private\_Models = Algorithm\_\ref{Algo IPDNNs}(training\_data)
\While{$\mathcal{D}$ has elements} \Comment{While stream has incoming data}
    \State Receive incoming data $x_t$ 
    \State pred, uncertainty $\leftarrow$ Private\_Models.predict($x_t$) 
    \State Add uncertainty to ADWIN 
    \If{ADWIN detects drift} 
        \State Request true labels $y_t$ for $x_t$ 
        \State Update training\_data with $x_t, y_t$ 
        \State Private\_Models = Algorithm\_\ref{Algo IPDNNs}(training\_data)
    \EndIf
\EndWhile
\end{algorithmic}
\end{algorithm}

Then any change detection algorithm such as ADWIN can be employed to detect drifts using this uncertainty measure. We chose ADWIN as it works well with real-valued inputs. The flowchart of the methodology is shown in Fig. \ref{flowchart}.  First $k$ integrally private models are computed from the initial available training data. With incoming data instances, we predict the output and input the prediction uncertainty from each of the k private models to ADWIN. If drift has been detected, true labels are requested and the training data must be updated with new instances. Here, our methodology does not require true labels to detect concept drift. New private models are computed from the updated training data. If the training data exceeds the threshold, records are removed on a first-come, first-served basis. In case there is no drift, then the prediction for new instances continues. 

Algorithm \ref{Algo IPDNNs} describes the computation for k private models. First, samples (i.e., sets of records) are generated from the training data in such a way that pairs of samples have empty intersection (i.e., they do not share any record). Models for these samples are computed using the same initialization. Models within $\Delta$ distance apart are kept in the same bucket. Buckets are sorted in descending order according to the number of models in each bucket. The algorithm then returns the mean of the top k recurring set of models as an ensemble of k integrally private models. 

Algorithm \ref{Algo IPDD} describes how to detect drifts privately. First, it uses initial available data as training data and computes private models using Algorithm \ref{Algo IPDNNs} on the training data. Predictions on new incoming instances are used to compute the uncertainty measure and to see if the drift is detected using ADWIN. If the drift is detected, true labels must be requested, then training data must be updated and the private models are recomputed on a new training data. This process continues as long as the new data is available for prediction. 

\subsection{Theoretical Analysis} \label{theoretical}

In this section, we present the probabilistic analysis for the recurrence of DNNs. DNNs are trained using mean samplers such as SGD, Adam etc. This analysis is inspired by the forgeability analysis done in \cite{thudi2022necessity}. The analysis in \cite{thudi2022necessity} computes the probability of forging a single minibatch while we focus on probabilistic analysis of learning the same model parameters after learning from different training data.  Let us consider a set of disjoint datasamples, $D_1,D_2,...,D_m$, i.i.d. (independent and identically distributed) sampled from a given $N$-dimensional dataset $\mathcal{D} $ with some distribution. Here, each of the $D_{i}$ is composed of $b$ minibatches $\hat{x} = {x_1, x_2 ,..., x_b}$. $M_1,M_2,...,M_m$ be the DNN models we want to train which have the same initialization. The update rule looks like $g(w, \hat{x}) = \frac{1}{b} \sum_{i=1}^b g(w, x_i)$. The update rule $g(w, x)$ can be seen as a random variable with mean $\mu$ and $\sigma^2$ ($ = \sum_{i=1}^N \sigma_i^2$, where $\sigma_i^2$ is the covariance of the $i^{th}$ component of a random variable $x$ sampled with distribution $\mathcal{D}$) as the trace of the covariance matrix. The mean sampler for the batch $\hat{x}, \; g(w, \hat{x})$ is still $\mu$ ($ \frac{1}{b} \sum_{i=1}^b g(w, x_i) = \frac{1}{b} * b\mu$) but individual variance will get the $\frac{1}{b}$ i.e. now the trace of the covariance matrix is $\frac{1}{b} \sigma^2$.

Since the data samples are i.i.d sampled from $\mathcal{D}$ and each $x_{i}$ is i.i.d sampled from  data samples, then $x_{i}$ follows the same distribution of $\mathcal{D}$. Then by Markov's inequality we can say that, 

\begin{align}
    P(|g(w, \hat{x}) - \mu|_2 \geq \Delta) = P(|g(w, \hat{x}) - \mu|_2^2 \geq \Delta^2) \leq \frac{E(|g(w, \hat{x}) - \mu|_2^2)}{\Delta ^2}
\end{align}

\begin{wrapfigure}{r}{0.36\textwidth}
    \includegraphics[width=0.36\textwidth]{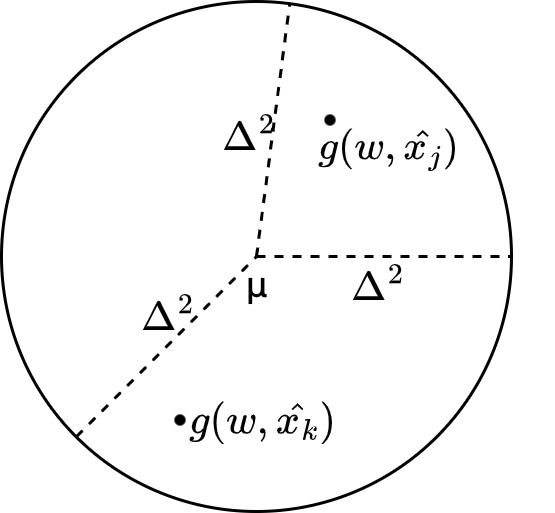}
    \caption{Two models $M_j, M_k$ at most $\Delta^2$ distance apart from $\mu$ with probability defined in Eq. (5)} \label{traiangle}
\end{wrapfigure}

Here the first equality is true by the property of monotonicity of squares and the second is Markov's inequality. Note that $E(|g(w, \hat{x}) - \mu|_2^2)$ is just the trace of the covariance matrix $(\frac{1}{b} \sigma^2)$. Then, we can write:

\begin{align}
    P(|g(w, \hat{x}) - \mu|_2^2 \geq \Delta^2) \leq \frac{\sigma ^2}{b\Delta ^2}
    \nonumber \\ \Rightarrow P(|g(w, \hat{x}) - \mu|_2^2 \leq \Delta^2) \geq 1 - \frac{\sigma ^2}{b\Delta ^2}
\end{align}

Let us consider two models $M_j \text{ and } M_k, $ training on data samples $D_j \text{ and }  D_k$ with $\hat{x_j}, \hat{x_k}$. From Eq. (5), we can say $ P(|g(w, \hat{x}_j) - \mu|_2^2 \leq \Delta^2) \geq (1 - \frac{\sigma ^2}{b\Delta ^2})$ and $P(g(w, \hat{x}_k) - \mu|_2^2 \leq \Delta^2) \geq (1 - \frac{\sigma ^2}{b\Delta ^2})$ at $i^{th}$ epoch

As demonstrated in Fig. \ref{traiangle}, if $g(w,\hat{x}_j), g(w,\hat{x}_k)$ are in the $\Delta^2$ ball of $\mu$ with probability defined in Eq. (5) then with probability $\geq (1-\frac{\sigma^2}{b\Delta^2})^2$ we can say both models are utmost $2\Delta^2$ distant. I.e. $P(|g(w,\hat{x}_j)-g(w,\hat{x}_k)| \leq 2 \Delta^2) \geq (1 - \frac{\sigma^2}{b\Delta^2})^2$. Then, the probability that out of $m$ models there exists two models which are in the $\Delta^2$ ball of $\mu$ would be:  
\begin{align}
    P(|g(w, \hat{x_j}) - g(w, \hat{x_k})|_2^2 \leq 2\Delta^2) \geq \sum_{r=2}^m \binom{m}{r} \left (1 - \frac{\sigma ^2}{b\Delta ^2} \right )^r \left (\frac{\sigma ^2}{b\Delta ^2}\right )^{m-r}
\end{align}

This is equivalent to having at least $2$ models out of $m$ in the $2\Delta^2$ ball of $\mu$. The probability that $m$ models are in the $2\Delta^2$ ball of $\mu$ would be:

\begin{align}
    P(|g(w, \hat{x_j}) - g(w, \hat{x_k})|_2^2 \leq 2\Delta^2) \geq \left (1 - \frac{\sigma ^2}{b\Delta ^2} \right)^m 
\end{align}

Equation (6) and (7) represents the probability of weights updating in the $2\Delta^2$ ball of $\mu$ for a single epoch. For $T$ iterations, the probability that there exists a model having weight updates in the $2\Delta^2$ ball of $\mu$ is at least $(\sum_{r=2}^m \binom{m}{r} (\frac{\sigma ^2}{b\Delta ^2})^{m-r} (1 - \frac{\sigma ^2}{b\Delta ^2})^r)^T$. After T epochs, the probability of $m$ models to be in the $2\Delta^2$ ball of $\mu$ will be at least $ ((1 - \frac{\sigma ^2}{b\Delta ^2})^m)^T$.

So, for samples sampled i.i.d. from some dataset, we can conclude that the lower bound for the probability that there exists recurrent models within $2\Delta^2$ ball is at least $ (\sum_{r=2}^m \binom{m}{r} (\frac{\sigma ^2}{b\Delta ^2})^{m-r} (1 - \frac{\sigma ^2}{b\Delta ^2})^r)^T$. In addition, the probability that all the m models are in the $2\Delta^2$ ball of $\mu$ is at least $ ((1 - \frac{\sigma ^2}{b\Delta ^2})^m)^T$. From this discussion, we have the following theorems.

\begin{theorem}

If $D_1, D_2, ..., D_m$ are i.i.d samples from the dataset $\mathcal{D} $ with some distribution and $b$ is the number of minibatches used for training in each of $T$ epochs. Then under similar training environment i.e. same initialization, learning rate, etc. with probability greater than $ (\sum_{r=2}^m \binom{m}{r} (\frac{\sigma ^2}{b\Delta ^2})^{m-r} (1 - \frac{\sigma ^2}{b\Delta ^2})^r)^T$, the model will recur. 

\end{theorem}

\begin{theorem}
    \vspace{1.5mm} With the above mentioned properties, a model satisfies k-anonymous integral privacy with probability atleast $(\sum_{r=k}^m \binom{m}{r} (\frac{\sigma ^2}{b\Delta ^2})^{m-r} (1 - \frac{\sigma ^2}{b\Delta ^2})^r)^T$
\end{theorem}
\textit{Proof:} See Eq. (6) for proof. k-Anonymity integral privacy is equivalent to having at least $k$ models out of $m$ in the $\Delta$ ball of $\mu$.



\textbf{Remark on the choice of $\Delta, m$:} In order to generate higher k-Anonymity integrally private models, from theorem 2 we can say that increasing the number of i.i.d samples $(m), \; b \; (\text{Number of batches used in each epochs}) \text{ and } \Delta $ (the distance value) increases the probability of getting recurrent models. 

\textbf{Role of initialization:} The probabilistic analysis presented here gives you the lower bound that the model will recur from the samples having similar distribution. The probability can further improve when models are initialized with the same weight as the learning from similar dataset would result in the similar learning for the models. 

\section{Experiments}

In this section, we present the experimental results for our proposed methodology. We will show that our methodology performs well with Categorical, Real, and Integer data with arbitrary number of classes. We perform our experiments on 3 real-world datasets namely Cover type (CovType), Electricity, and Susy dataset \cite{Dua:2019}. We also run our experiments on artificially generated Sine data and Insects data with abrupt, gradual and incremental drifts \cite{souza2020challenges}. Table \ref{tab:used dataset} shows the number of instances and other details of these datasets. 

\begin{table}[]
    \centering
    \begin{tabular}{|c|c|c|c|c|}
        \hline
        Dataset & \# instances & \# attribute & Data type & \# classes \\
        \hline
        CovType & 581012 & 54 & \makecell{Categorical \\ Integer} & 7 \\
        \hline
        Electricity & 45312 & 8 & \makecell{Real \\ Integer} & 2 \\
        \hline
        Susy & 5000000 & 18 & Real & 2 \\
        \hline
        Sine & 200000 & 4 & Real & 2 \\
        \hline
        Insects\_ab & 52848 & 33 & Real & 6 \\
        \hline
        Insects\_grad & 24150 & 33 & Real & 6 \\
        \hline
        Insects\_incre & 57018 & 33 & Real & 6 \\
        \hline
    \end{tabular}
    \caption{Details of the used Datasets}
    \label{tab:used dataset}
\end{table}

For our experiments, we have randomly considered a NN with a single hidden layer (10 neurons) architecture (We will call this architecture ANN) and a three hidden layer NN architecture with 10-20-10 number of neurons (we will call it DNN). For our experimental purpose we have chosen $\Delta=0.01$ and ADWIN parameter, $\delta=0.001$. For all the datasets, we have initially trained ANN and DNN over 10\% of the dataset, and then stream is evaluated with 2\% of the dataset at each time instance. 

We compare  our results (Integrally private drift detection, IPDD) with No re-training (No\_retrain), ADWIN with unlimited label availability (ADWIN\_unlim), and ADWIN with limited labels (ADWIN\_lim). We have used three levels of differentially private models: high privacy ($\epsilon = 0.1$) under limited label availability (DP\_01), moderate privacy ($\epsilon = 0.5$) under limited label availability (DP\_05) and low privacy ($\epsilon = 1.0$) under limited label availability (DP\_10). All the results have been computed for ANN as well as DNN. The No\_retraining model approach does not check for drifts, it trains the model with initial data once and only does the prediction for the rest of the data stream. For ADWIN\_unlim we assume it has access to all the true labels of the incoming data stream and it detects drifts using the true labels only. The ADWIN\_lim can have true labels upon request but detect the drifts using uncertainty through the ADWIN model. Similar settings were assumed for DP\_01, DP\_05, DP\_10 and IPDD. 

We can observe that our methodology IPDD has better or comparable accuracy score for both ANN and DNN. Table \ref{tab:acc_score} provides the accuracy of the learned models. IPDD performs better than its counterparts for CovType and Electricity datasets, it has comparable accuracy score for Insects\_ab and comparable results with ADWIN\_unlim method. Table \ref{tab:mcc_score} provides the results for Mathews correlation coefficient (mcc) in the range $[-1, 1]$ (higher the better). MCC is a reliable statistical rate which assigns high value to a classifier if it performs good in all four confusion matrix categories. In comparison with differentially private models, IPDD performs much better than all three levels of differential privacy for all datasets except Insects\_grad and Insects\_incre. For ANN, IPDD performs performs better for CovType dataset and has comparable mcc rate with the rest. In case of DNNs, IPDD performs better than its counterparts for Electricity, Susy and Insects\_ab datasets; and performs comparable results for the rest of the datasets. Table \ref{tab:auc_score} shows the score for the area under the curve (auc score). Auc score is the probability that a model ranks a random positive instance higher than a random negative instance. Table \ref{tab:auc_score} highlights that auc score for IPDD's ANN and DNN performs better than its counterparts in case of all the datasets except Insect\_grad and Insect\_incre dataset.

\begin{wrapfigure}{h}{0.36\textwidth}
    \includegraphics[width=0.36\textwidth]{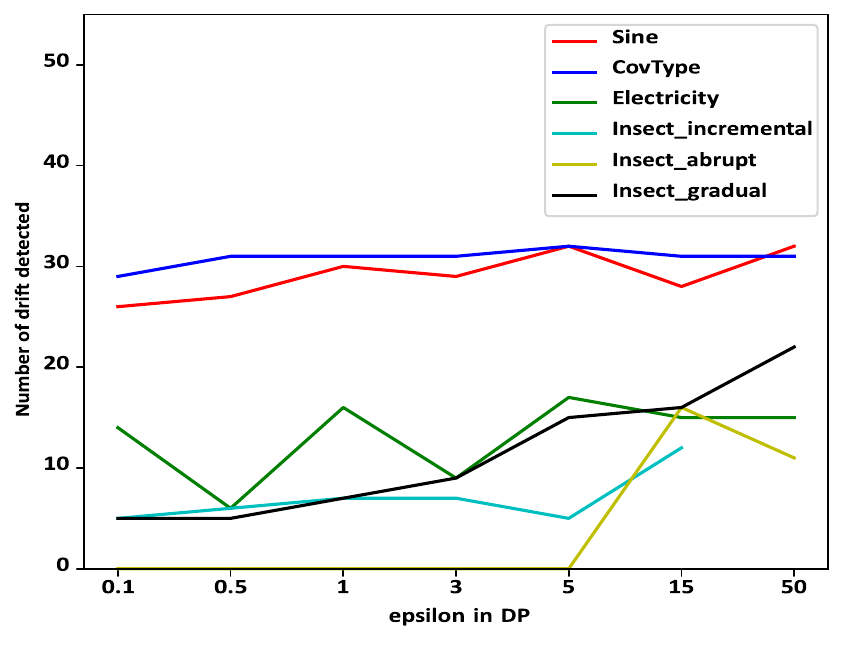}
    \caption{Drift detected by different $\epsilon$-differentially private models} \label{drift_detected DP}
\end{wrapfigure}

We observed that with the addition of noise DP models may struggle to detect drifts. Table \ref{tab:drift_detected} shows the number of drifts detected by each method. It highlights that DP models at times may detect very few drifts because of the noise. On the other hand, IPDD detects comparable drifts to ADWIN\_unlim and ADWIN\_lim for both ANN and DNN. Table \ref{tab:drift_detected} also highlights that proposed IPDD does not detect unnecessary drifts i.e. IPDD does not necessary detect any drift when there is none (counterparts of IPDD does not detect any drift). As expected when the noise for DP models decreases, more drifts were detected by DP models as shown in Fig. \ref{drift_detected DP}.

\begin{figure}
\centering
    \begin{subfigure}[b]{0.24\textwidth}
         \centering
         \includegraphics[width=\textwidth]{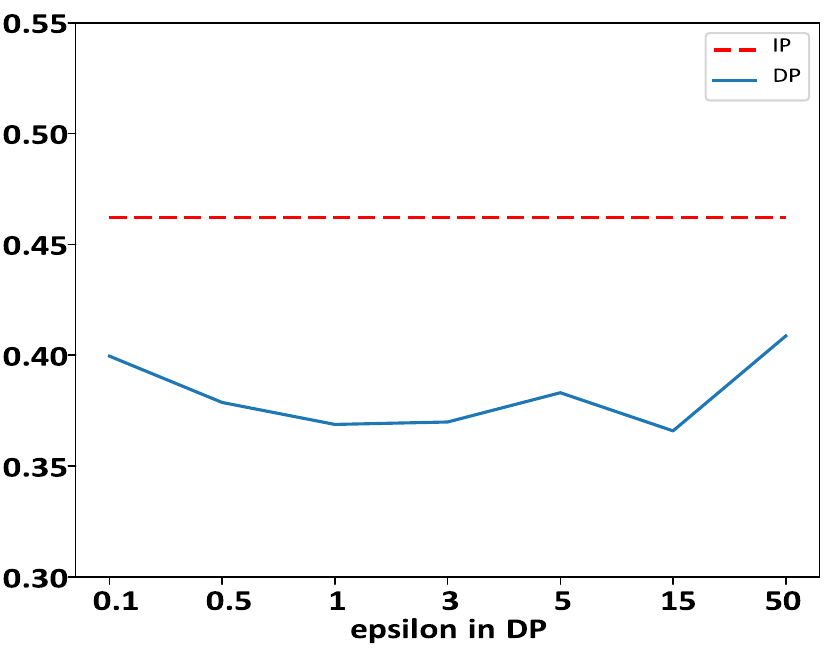}
         \caption{ }
         \label{covtype_DPvsIPDD}
     \end{subfigure}
     \hfill
     \begin{subfigure}[b]{0.24\textwidth}
         \centering
         \includegraphics[width=\textwidth]{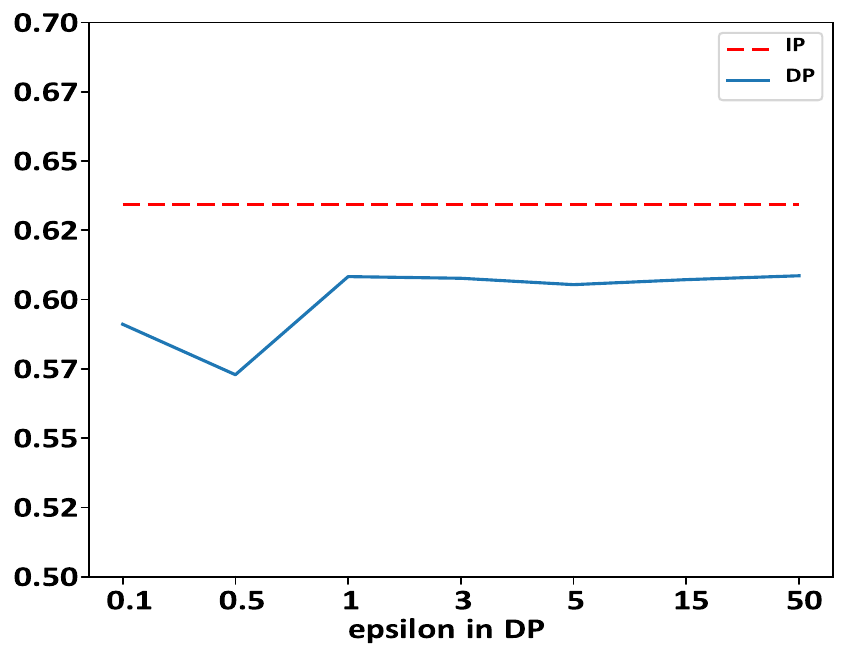}
         \caption{ }
         \label{elec DPvsIPDD}
     \end{subfigure}
     \hfill
     \begin{subfigure}[b]{0.24\textwidth}
         \centering
         \includegraphics[width=\textwidth]{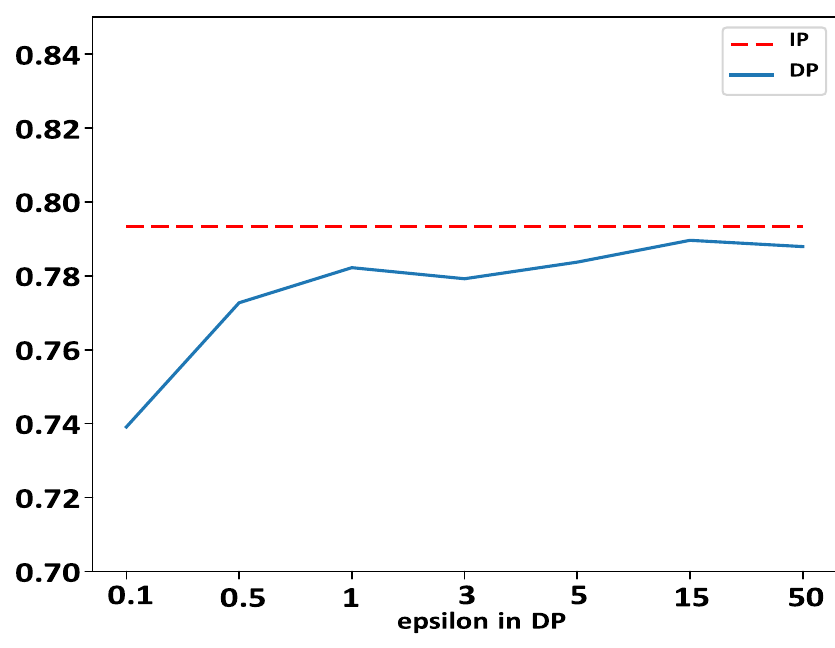}
         \caption{ }
         \label{susy DP vsIPDD}
     \end{subfigure}
     \hfill
     \begin{subfigure}[b]{0.24\textwidth}
         \centering
         \includegraphics[width=\textwidth]{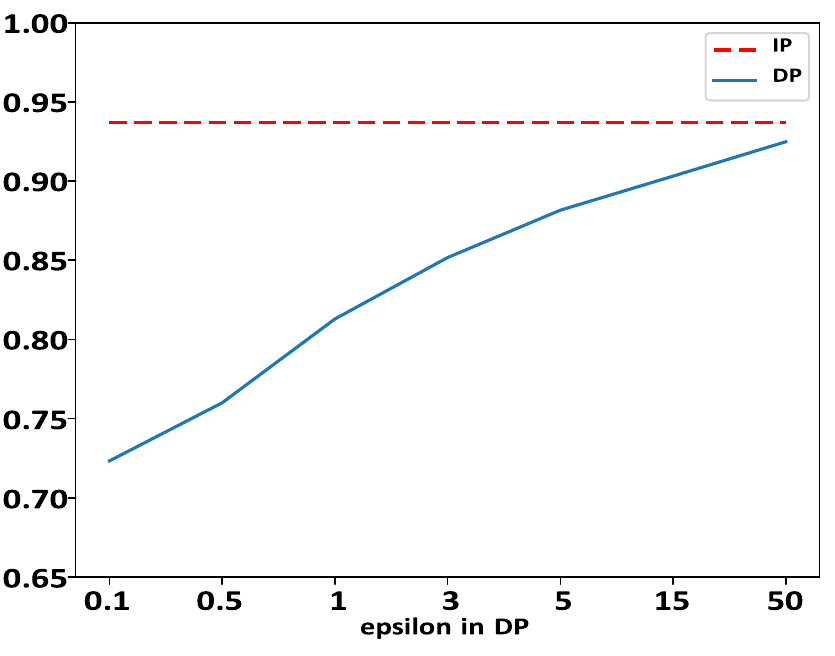}
         \caption{ }
         \label{sine dpvsIPDD}
     \end{subfigure}
     \hfill
     \begin{subfigure}[b]{0.24\textwidth}
         \centering
         \includegraphics[width=\textwidth]{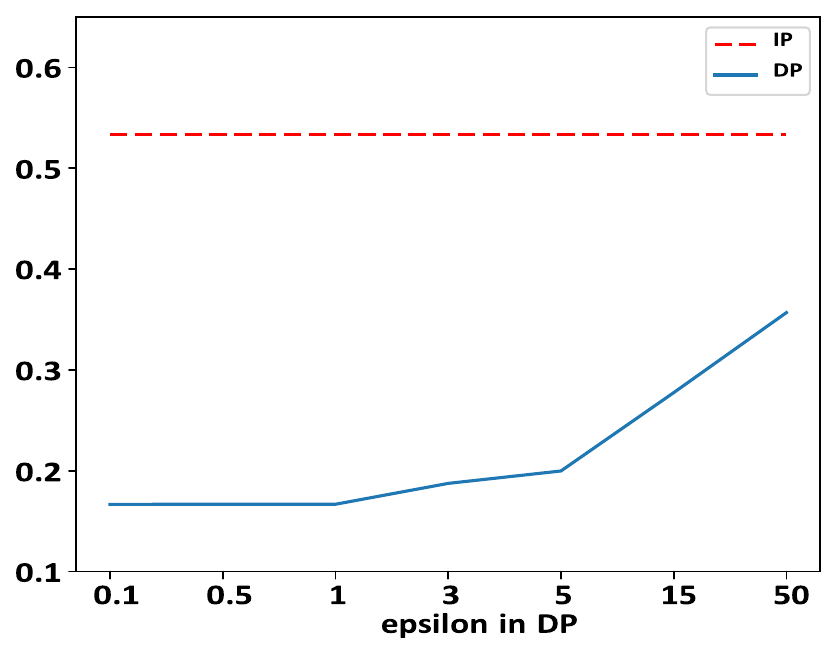}
         \caption{ }
         \label{insect_ab dpvsipdd}
     \end{subfigure}
     \hfill
     \begin{subfigure}[b]{0.24\textwidth}
         \centering
         \includegraphics[width=\textwidth]{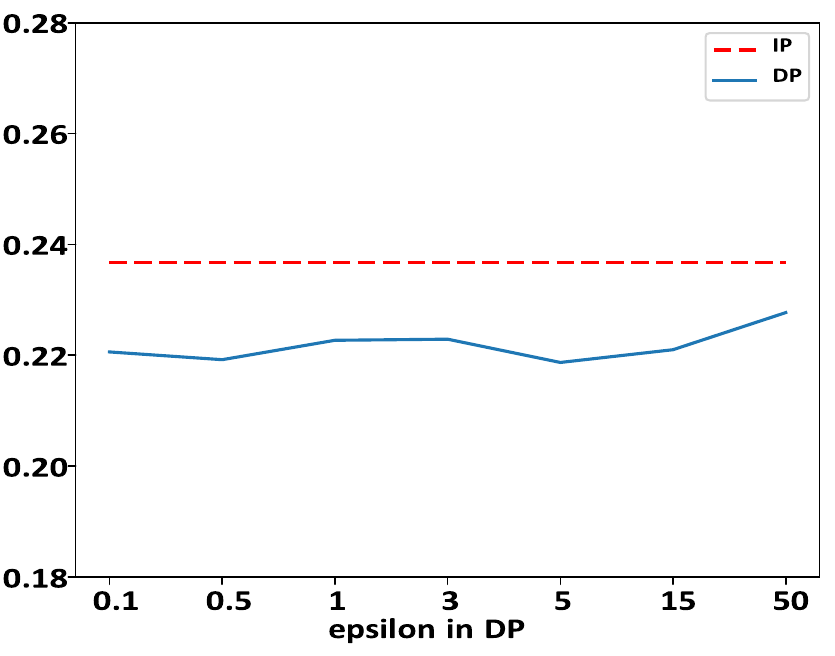}
         \caption{ }
         \label{insect_grad dpvsipdd}
     \end{subfigure}
     \hfill
     \begin{subfigure}[b]{0.24\textwidth}
         \centering
         \includegraphics[width=\textwidth]{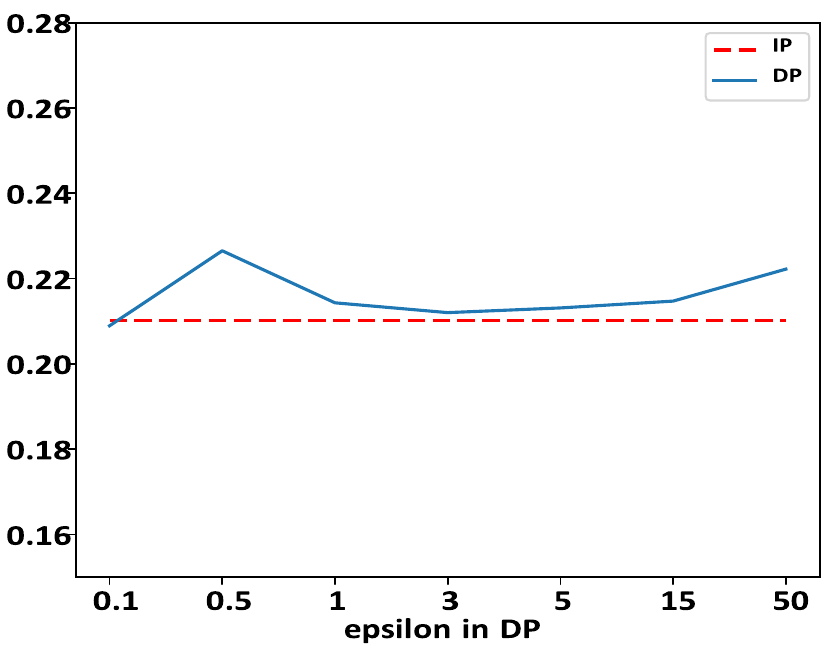}
         \caption{ }
         \label{insect_incr dpvsipdd}
     \end{subfigure}
     
\caption{ Comparison of the accuracy score between differential privacy and integral privacy: (a) CovType (b) Electricity (c) Susy (d) Sine (e) Insect\_ab (f) Insect\_grad (g) Insect\_incr.}
\label{Compare Acc DPvsIPDD}
\end{figure}

In most of the cases, differentially private models does not perform as good as IPDD models even when the $\epsilon$ is very high (very low privacy). This is shown in Fig. \ref{Compare Acc DPvsIPDD}. It compares the accuracy score between DNN model of DP and IPDD for all the datasets. Fig. \ref{covtype_DPvsIPDD}, \ref{elec DPvsIPDD}, \ref{susy DP vsIPDD}, \ref{sine dpvsIPDD}, \ref{insect_ab dpvsipdd}, and \ref{insect_grad dpvsipdd} highlights that even though the accuracy score improves for DP models, IPDD still performs better than DP. Only in case of Fig. \ref{insect_incr dpvsipdd} the DP model has slightly better accuracy score than IPDD even in case of high privacy.

\afterpage{\clearpage 
\begin{landscape}
    \centering
        \captionof{table}{Accuracy Score.}\label{tab:acc_score}\par
        \begin{tabular}{| *{15}{c|}}
            \hline
            Dataset & \multicolumn{2}{c|}{No\_retrain} & \multicolumn{2}{c|}{Adwin\_unlim} & \multicolumn{2}{c|}{Adwin\_lim} & \multicolumn{2}{c|}{DP\_01} & \multicolumn{2}{c|}{DP\_05} & \multicolumn{2}{c|}{DP\_10} & \multicolumn{2}{c|}{IPDD} \\
            \hline
              & ANN & DNN & ANN & DNN & ANN & DNN & ANN & DNN & ANN & DNN & ANN & DNN & ANN & DNN \\
            \hline
            CovType & 0.4182 & 0.3552 & 0.4437 & 0.4582 & 0.4246 & 0.4012 & 0.4789 & 0.3996 & 0.4434 & 0.3787 & 0.4563 & 3688 & \textbf{0.5836} & \textbf{0.4621} \\
            \hline
            Electricity	& 0.5754 & 0.5790 & 0.5928 & 0.6154 & 0.5953 & 0.6096 & 0.5875 & 0.5911 & 0.5994 & 0.5729 & 0.5978 & 0.6083 & \textbf{0.6281} & \textbf{0.6343} \\
            \hline
            Susy & \textbf{0.7985} & \textbf{0.7985} & \textbf{0.7985} & \textbf{0.7985} & \textbf{0.7985} & \textbf{0.7985} & 0.6844 & 0.7391 & 0.7327 & 0.7727 & 0.7539 & 0.7822 & 0.7729 & \textbf{0.7934} \\
            \hline
            Sine & \textbf{0.9505} & \textbf{0.9536} & 0.93425 & 0.9677 & 0.9421 & 0.9577 & 0.6204 & 0.6134 & 0.7600 & 0.6447 & 0.8129 & 0.9005 & \textbf{0.9365} & \textbf{0.9369} \\
            \hline
            Insects\_ab & 0.4416 & 0.4434 & 0.5252 & 0.5237 & 0.2903 & 0.5348 & 0.1739 & 0.1666 & 0.2437 & 0.1667 & 0.2126 & 0.1667 & \textbf{0.5241} & \textbf{0.5331} \\
            \hline
            Insects\_grad & \textbf{0.2810} & \textbf{0.2841} & 0.2293 & 0.2412 & 0.2221 & 0.232 & 0.2104 & 0.2206 & 0.2238 & 0.2192 & 0.2213 & 0.2224 & 0.2227 & 0.2368\\
            \hline
            Insects\_incr & 0.2181 & \textbf{0.2475} & 0.2282 & \textbf{0.2475} & 0.2335 & 0.2437 & 0.2140 & 0.2089 & 0.1914 & 0.2255 & 0.2143 & 0.2143 & 0.2022 & 0.2101 \\
            \hline
        \end{tabular}
    \captionof{table}{Mathews Correlation Coefficient.}\label{tab:mcc_score}\par
        \begin{tabular}{| *{15}{c|}}
            \hline
            Dataset & \multicolumn{2}{c|}{No\_retrain} & \multicolumn{2}{c|}{Adwin\_unlim} & \multicolumn{2}{c|}{Adwin\_lim} & \multicolumn{2}{c|}{DP\_01} & \multicolumn{2}{c|}{DP\_05} & \multicolumn{2}{c|}{DP\_10} & \multicolumn{2}{c|}{IPDD} \\
            \hline
              & ANN & DNN & ANN & DNN & ANN & DNN & ANN & DNN & ANN & DNN & ANN & DNN & ANN & DNN \\
            \hline
            CovType & 0.0816 & 0.1940 & 0.1618 & 0.2941 & 0.1421 & 0.2140 & 0.1012 & 0.0599 & 0.0183 & 0.0261 & -0.0158 & \textbf{-0.0014} & \textbf{0.3575} & \textbf{0.2871} \\
            \hline
            Electricity & 0.2652 & 0.0 & 0.1459 & 0.2101 & 0.1559 & 0.1898 & 0.0906 & 0.1120 & 0.1615 & 0.0033 & 0.1576 & 0.1817 & \textbf{0.2593} & \textbf{0.2837} \\
            \hline
            Susy & \textbf{0.5936} & \textbf{0.5936} & \textbf{0.5936} & \textbf{0.5936} & \textbf{0.5936} & \textbf{0.5936} & 0.3608 & 0.4782 & 0.4627 & 0.5411 & 0.5050 & 0.5623 & 0.5591 & \textbf{0.5902} \\
            \hline
            Sine & 0.8995 & 0.9059 & 0.8667 & \textbf{0.9345} & 0.8824 & 0.9142 & 0.1879 & 0.1974 & 0.5105 & 0.3335 & 0.6670 & 0.8024 & \textbf{0.8726} & \textbf{0.8722} \\  
            \hline
            Insects\_ab & 0.3841 & 0.3849 & \textbf{0.4412} & \textbf{0.4398} & 0.1274 & 0.4528 & 0.0129 & 0.0 & 0.1164 & 0.0014 & 0.0588 & 0.0 & \textbf{0.4397} & \textbf{0.4505} \\
            \hline
            Insects\_grad & 0.1356 & 0.1181 & 0.0967 & \textbf{0.1586} & 0.0858 & 0.1291 & 0.0675 & 0.1176 & 0.1100 & 0.1333 & 0.1248 & 0.1258 & 0.1112 & 0.1181 \\
            \hline
            Insects\_incr & 0.0628 & \textbf{0.0978} & 0.0746 & \textbf{0.0978} & 0.0809 & 0.0962 & 0.0834 & 0.0755 & 0.0356 & 0.0924 & 0.0902 & 0.0954 & 0.0448 & 0.0553 \\
            \hline
        \end{tabular}
\end{landscape}  
}

\afterpage{\clearpage
\begin{landscape}
    \centering
    \captionof{table}{Auc Score.}\label{tab:auc_score}\par
        \begin{tabular}{| *{15}{c|}}
            \hline
            Dataset & \multicolumn{2}{c|}{No\_retrain} & \multicolumn{2}{c|}{Adwin\_unlim} & \multicolumn{2}{c|}{Adwin\_lim} & \multicolumn{2}{c|}{DP\_01} & \multicolumn{2}{c|}{DP\_05} & \multicolumn{2}{c|}{DP\_10} & \multicolumn{2}{c|}{IPDD} \\
            \hline
              & ANN & DNN & ANN & DNN & ANN & DNN & ANN & DNN & ANN & DNN & ANN & DNN & ANN & DNN \\
            \hline
            CovType & 0.7349 & 0.8801 & 0.7985 & \textbf{0.9058} & 0.7982 & 0.8596 & 0.5188 & 0.5197 & 0.5679 & 0.5422 & 0.5910 & 0.5293 & \textbf{0.8881} & \textbf{0.9087} \\
            \hline
            Electricity & 0.6195 & 0.5 & 0.5209 & 0.5508 & 0.5239 & 0.5433 & 0.5261 & 0.5430 & 0.5299 & 0.5004 & 0.5277 & 0.5424 & \textbf{0.6311} & \textbf{0.6431} \\ 
            \hline
            Susy & \textbf{0.7923} & \textbf{0.7923} & \textbf{0.7923} & \textbf{0.7923} & \textbf{0.7923} & \textbf{0.7923} & 0.6751 & 0.7275 & 0.7217 & 0.7674 & 0.7451 & 0.7741 & \textbf{0.7582} & \textbf{0.7828} \\
            \hline
            Sine & 0.9492 & 0.9531 & 0.9311 & \textbf{0.9677} & 0.9403 & 0.9565 & 0.5769 & 0.5776 & 0.7247 & 0.5979 & 0.8156 & 0.8912 & \textbf{0.9381} & \textbf{0.9339} \\
            \hline
            Insects\_ab & 0.7989 & 0.80 & 0.8469 & 0.8587 & 0.6310 & 0.8469 & 0.5829 & 0.5137 & 0.5966 & 0.5749 & 0.5528 & 0.5516 & \textbf{0.8474} & \textbf{0.8581} \\
            \hline
            Insects\_grad & {0.6283} & 0.6206 & 0.6190 & 0.6068 & 0.6024 & 0.6118 & 0.5377 & \textbf{0.6220} & 0.5800 & 0.6067 & 0.5779 & 0.6219 & 0.5508 & \textbf{0.5956} \\
            \hline
            Insects\_incr & 0.5439 & 0.5795 & 0.5470 & 0.5799 & 0.5577 & 0.5730 & 0.6026 & 0.5841 & 0.6035 & \textbf{0.6245} & 0.6044 & 0.6161 & 0.5249 & 0.5362 \\
            \hline
        \end{tabular}
    \captionof{table}{Number of Drifts detected by each algorithm.}\label{tab:drift_detected}\par
    \begin{tabular}{| *{15}{c|}}
        \hline
        Dataset & \multicolumn{2}{c|}{No\_retrain} & \multicolumn{2}{c|}{Adwin\_unlim} & \multicolumn{2}{c|}{Adwin\_lim} & \multicolumn{2}{c|}{DP\_01} & \multicolumn{2}{c|}{DP\_05} & \multicolumn{2}{c|}{DP\_10} & \multicolumn{2}{c|}{IPDD} \\
        \hline
          & ANN & DNN & ANN & DNN & ANN & DNN & ANN & DNN & ANN & DNN & ANN & DNN & ANN & DNN \\
        \hline
        CovType & 0 & 0 & 35 & 35 & 31 & 32 & 24 & 29 & 24 & 31 & 18 & 31 & \textbf{35} & \textbf{33} \\
        \hline
        Electricity & 0 & 0 & 37 & 37 & 42 & 39 & 13 & 14 & 8 & 6 & 15 & 16 & \textbf{23} & \textbf{24} \\
        \hline
        Susy & 0 & 0 & 0 & 0 & 0 & 0 & 0 & 0 & 0 & 0 & 0 & 0 & 0 & 0 \\
        \hline
        Sine & 0 & 0 & 28 & 28 & 33 & 28 & 27 & 31 & 27 & 21 & 28 & 28 & 17 & 14 \\
        \hline
        Insects\_ab & 0 & 0 & 24 & 24 & 23 & 17 & \textbf{0} & \textbf{0} & \textbf{5} & \textbf{0} & \textbf{12} & \textbf{0} & 16 & 14 \\
        \hline
        Insects\_grad & 0 & 0 & 18 & 18 & 16 & 17 & \textbf{5} & \textbf{5} & \textbf{5} & \textbf{5} & \textbf{5} & \textbf{7} & 14 & 15 \\
        \hline
        Insects\_incr & 0 & 0 & 0 & 0 & 16 & 14 & \textbf{7} & \textbf{5} & \textbf{8} & \textbf{6} & \textbf{6} & \textbf{7} & \textbf{15} & \textbf{19} \\
        \hline
    \end{tabular}
\end{landscape}
}

\begin{figure}
\centering
    \begin{subfigure}[b]{0.45\textwidth}
         \centering
         \includegraphics[width=\textwidth]{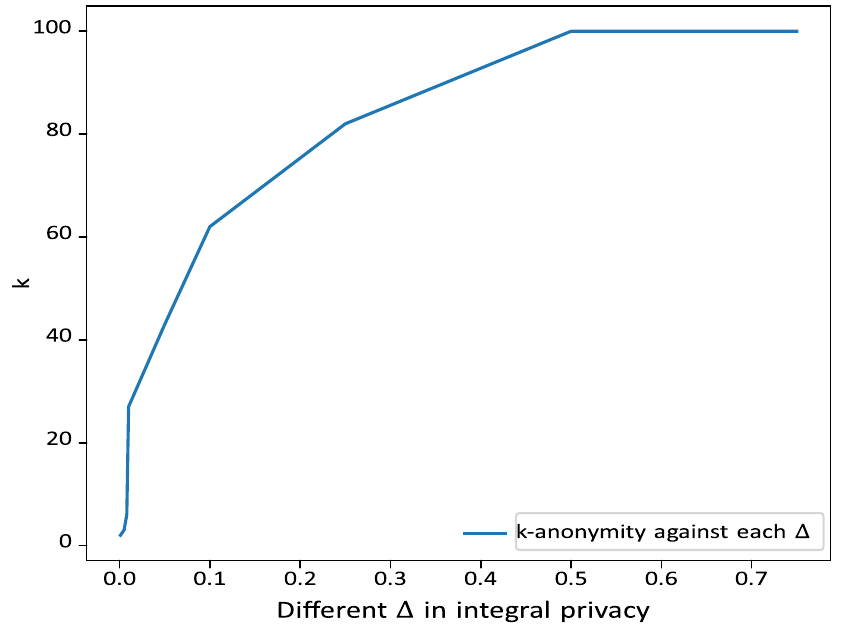}
         \caption{ }
         \label{increasing epsilon}
     \end{subfigure}
     \hfill
     \begin{subfigure}[b]{0.45\textwidth}
         \centering
         \includegraphics[width=\textwidth]{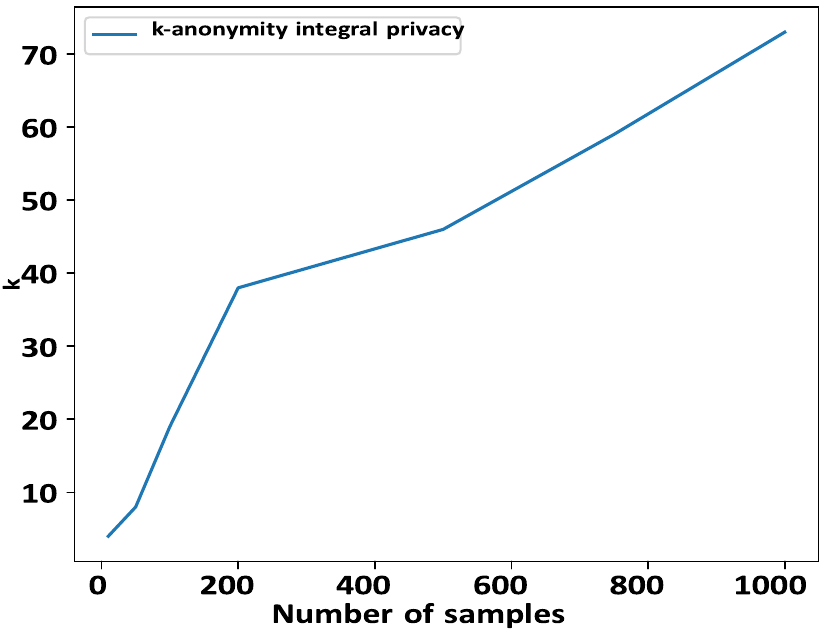}
         \caption{ }
         \label{increasing samples}
     \end{subfigure}
\caption{K-anonymity integral privacy against (a) increasing $\Delta$ (b) increasing the number of i.i.d samples}
\label{k-anonymity ip}
\end{figure} 

\begin{figure}
\centering
    \begin{subfigure}[b]{0.30\textwidth}
         \centering
         \includegraphics[width=\textwidth]{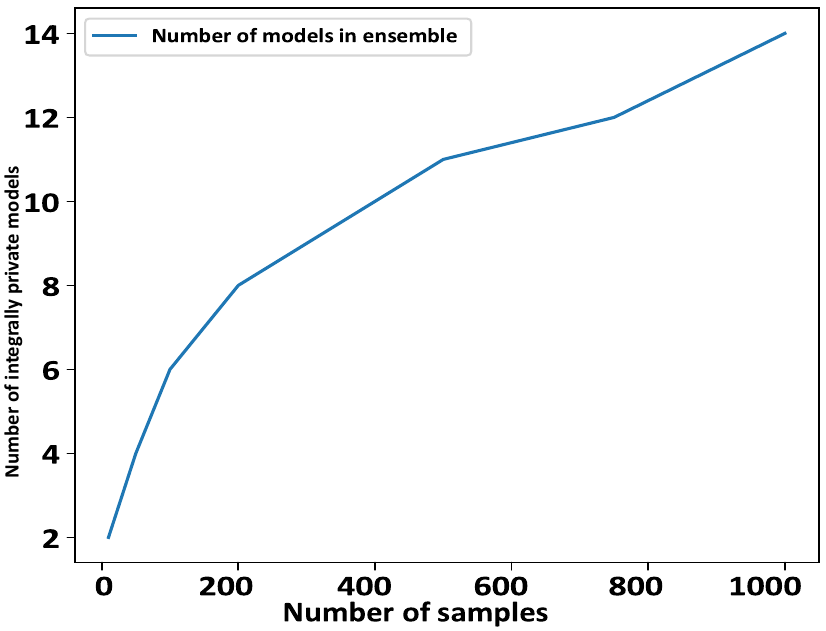}
         \caption{ }
         \label{increasing samples k-anonymous}
     \end{subfigure}
     \hfill
     \begin{subfigure}[b]{0.30\textwidth}
         \centering
         \includegraphics[width=\textwidth]{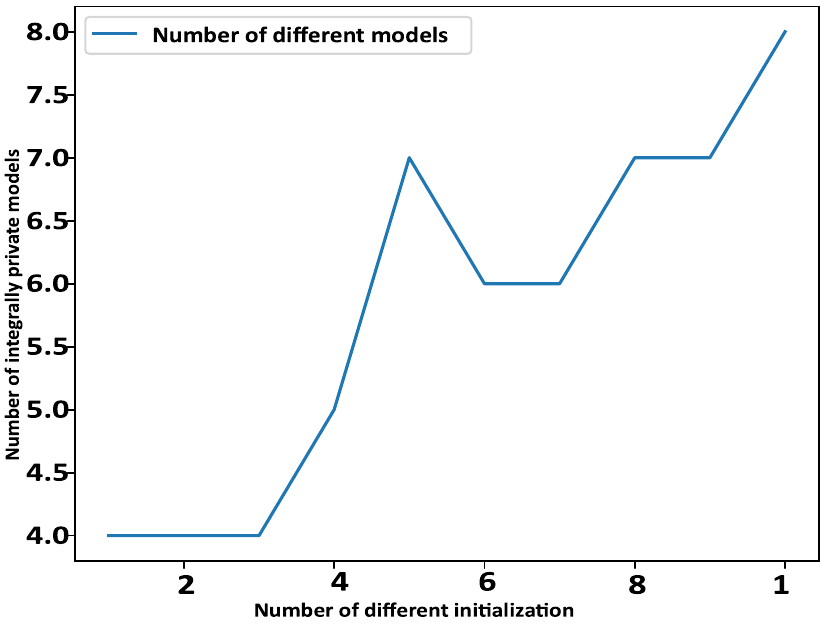}
         \caption{ }
         \label{different initializations}
     \end{subfigure}
     \hfill
     \begin{subfigure}[b]{0.30\textwidth}
         \centering
         \includegraphics[width=\textwidth]{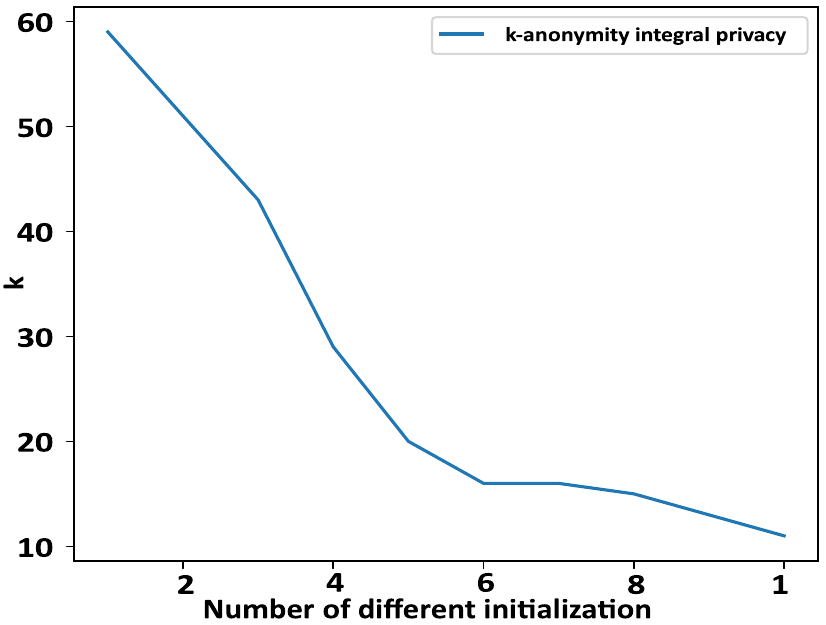}
         \caption{ }
         \label{different initializations k-anonymity}
     \end{subfigure}
\caption{ Number of integral private models in an ensemble against (a) increasing the number of i.i.d samples (b) increasing the number of different initialization. (c) k-anonymity against different initialization}
\label{k-anonymous IP}
\end{figure}

Section \ref{theoretical} shows the probabilistic analysis on the lower bound of the recurrence of Integrally private models. As discussed in the remarks of Section \ref{theoretical}, the higher the value of $\Delta$ the higher the number k-anonymity in integrally private models. For models with same initialization trained on 100 i.i.d samples ($D_1,D_2,...,D_{100}$) from Sine dataset, the k in k-anonymity Integral privacy has been plotted against increasing $\Delta$ in Fig. \ref{increasing epsilon}. As can be seen increasing the $\Delta$ value leads to the higher value of k in k-anonymity Integral privacy. Similarly, we can see in Fig. \ref{increasing samples} that for a fixed $\Delta = 0.01$, increasing the number of i.i.d samples leads to a higher value of k in k-Anonymity integral privacy. It is important to highlight the distinction between k-anonymity and ensemble of k models chosen with k-anonymity Integral privacy. Simply, in k-Anonymity integral privacy, there exists a bucket which has at least k models while we require an ensemble of k such buckets.

We observe in Fig. \ref{increasing samples k-anonymous} that for a fixed $\Delta = 0.01$, increasing the number of i.i.d samples also leads to the higher k in k-Anonymity integral privacy. In cases where all the models are clustered to only one IP model, generating an ensemble of such models can be tricky. An easier way to avoid this problem is to generate an ensemble of k-anonymity models using different initializations. The reason for this could be attributed to the comparable learning process when using similar training data. For 100 i.i.d samples of Sine data, in Fig. \ref{different initializations}, x-axis shows the number of different initialization and y-axis shows the number of different IP models in an ensemble. It is important to note here that for 100 samples if the number of IP models increases in an ensemble, k-anonymity of each IP model will decrease as depicted in Fig. \ref{different initializations k-anonymity}. 
 
\subsection{Limitations of our approach:} The analysis of our method as well as our experiment permits us to state the following.

\begin{enumerate}
    \item Generating k-anonymous integrally private models requires training on large number of samples which is a time consuming process. The proposed IPDD methodology has running time as the cost of privacy. 
    \item As shown in Section \ref{theoretical}, the generation of integrally private models is a probabilistic approach and depends on the samples selected. That is, different runs can provide different results. 
\end{enumerate}

\section{Conclusion and Future work}

In this paper we have presented a private drift detection methodology called 'Integral Privacy Drift Detection' (IPDD). Our methodology detects drifts using an ensemble of k-anonymity integrally private models. Simply, we generate an ensemble of k models which are recurring from multiple disjoint datasets. Our methodology does not require the ground truth to detect concept drift but assumes they are available for retraining. We find that our methodology can successfully detect concept drifts while maintaining the utility of non-private models.  It is useful in generating models which have comparable (better in some cases) accuracy score, mcc score and auc score against ADWIN with unlimited label availability and limited label availability. In comparison with its differentially private counterpart, IPDD perfoms significantly better in most of the cases.

As shown above different parameters can lead to different levels of privacy. It can also affect the number of drifts detected and the utilty of the model. Fine-tuning of these parameters for each application is an interesting direction for future work. Furthermore, extension of our work for non-i.i.d. samples would be an interesting future direction.

\vspace{0.2cm}

\textbf{Acknowledgement: } This work was partially supported by the Wallenberg Al, Autonomous Systems and Software Program (WASP) funded by the Knut and Alice Wallenberg Foundation. The computations were enabled by the supercomputing resource Berzelius provided by National Supercomputer Centre at Linköping University and the Knut and Alice Wallenberg foundation.

\printbibliography
\end{document}